\documentclass{article}

\usepackage{PRIMEarxiv}

\usepackage[utf8]{inputenc} 
\usepackage[T1]{fontenc}    
\usepackage{hyperref}       
\usepackage{url}            
\usepackage{booktabs}       
\usepackage{amsfonts}       
\usepackage{nicefrac}       
\usepackage{microtype}      
\usepackage{lipsum}
\usepackage{fancyhdr}       
\usepackage{graphicx}       
\graphicspath{{images/}}     
\usepackage{multicol}
\usepackage[export]{adjustbox}

\title{A degree of image identification at sub-human scales could be possible with more advanced clusters
}

\author{
  Prateek Y J \\
  Sponsored by Microsoft India\\
  \texttt{prateekyashwant.jannu2020@vitstudent.ac.in} \\
}

\begin{document}
\maketitle

\begin{abstract}
The purpose of the research is to determine if currently available self-supervised learning techniques can accomplish human level comprehension of visual images using the same degree and amount of sensory input that people acquire from. Initial research on this topic solely considered data volume scaling. Here, we scale both the volume of data and the quality of the image.
This scaling experiment is a self-supervised learning method that may be done without any outside financing.We find that scaling up data volume and picture resolution at the same time enables human-level item detection performance at sub-human sizes.We run a scaling experiment with vision transformers trained on up to 200000 images up to 256 ppi.

\end{abstract}

\section{Introduction}
While initial investigations in this domain primarily focused on the scaling of data volume, the present research takes a bold leap by not only scaling the data volume but also enhancing the image quality. This ambitious scaling experiment adopts a self-supervised learning approach that is both resource-efficient and feasible even without external financial support. Remarkably, our findings unveil the possibility of achieving human-level image identification performance at scales below human capabilities, achieved through the simultaneous scaling of data volume concentration and snapshot resolution. In order to do this, we conduct a scaling experiment using vision transformers and train them on a large data set with up to 200000 pictures, each of which has a resolution of 256 ppi.

\section{Evaluation Factors and Issues}
I answer that query in relation to a particular skill, namely the capacity to recognize visual images in the real world. It is difficult to directly assess the data efficiency of deep learning models trained with self supervised learning methods with regard to people for a variety of explanations.\\
\subsection{Inconsistency in dimensions}Inconsistency in dimensions: They model frequently function with significantly smaller impulses than the human cerebral cortex does, as seen, for instance, when comparing the size of common images used in computer vision with the quantity of detectors in the parafovea of an average person.

\subsection{Inconsistency in Model Scale}
 When assessing the number of parameters in a model to the number of neurotransmitters in the functioning model they are often significantly less than the functioning nervous system, or indeed  the visual regions within the nervous system.

\section{Control Conditions}
\label{sec:headings}
We undertake control conditions to find out whether the present Self-Learning algorithms can reach sub-human image cognizant scales for model scale, and image dimensions in order to address these discrepancies. In this section we take into account scaling both components simultaneously and come to a decision.
In the present trials, we additionally train our models on a roughly two-fold bigger collection of images that resemble real beings.

\subsection{Experimental data}
A total of two-thousand images of human-like footage from four different data-sets make up the whole collection of training data. The videos have two distinct characteristics: In contrast to common pictures data-sets in machine learning, which usually consist of significantly fewer photographs, most of the content are (i) organic, independent head-cam videos taken from the perspective of grown-up or toddler camera users as they go about their daily lives, and (ii) these are time-wise extended, continuous videos with typical run-times of tens of a few seconds to minutes. The following clip data-sets make up the merged training set: \\
The individual picture data-sets that make up our integrated training data set are broken out in Figure 1 along with their sizes in thousands. Image-Net, which makes over 50\% of our total training data, is by far the largest contributor. About 31.5\% of the training data come from the CelebA dataset, while 5\% come from the CIFAR-10 dataset and 13.5\% come from the ADE20K dataset. As a consequence, we anticipate that people who choose to repeat the experiments in this study using solely publicly available data sources would obtain outcomes that are strikingly comparable to those presented here.

We train models on the whole dataset and on constantly randomized random portions of it in order to investigate the scaling of image identification performance with the quantity of human-like visual input utilized during self-supervised pretraining. We explicitly construct models on percentages of the dataset that are 100 percent, 50 percent, 25 percent and 5 percent. These selections encompass a 20-fold variance in data size, encompassing around two hundred thousand to ten thousand captures of labelled images in every instance (from the biggest through the tiniest). We repeat the subset selection a total of four times for every data set size because it is stochastic.

\subsubsection{Image-Net}
The Image-Net Large Scale Visual Recognition Challenge (ILSVRC) 2012–2017 image categorization and localization data-set is the most frequently used subset of Image-Net\cite{russakovsky2015imagenet}. There are 1,281,167 training photos, 50,000 validation images, and 100,000 test images in this dataset, which covers 1000 item classes. On Kaggle, this subset is accessible \href{https://www.image-net.org/download.php}{here}.
\begin{figure}
    \centering
    \includegraphics[width=10cm]{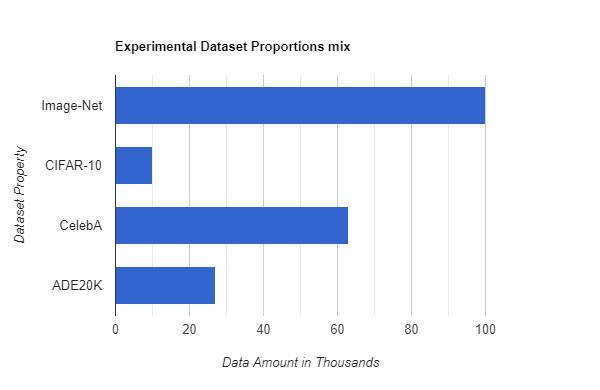}
    \caption{Image Data-set proportions used for control-Conditions}
    \label{fig:fig1}
\end{figure}
\subsubsection{CIFAR-10}
Five learning groups and one evaluation batch, each containing 10,000 snapshots, make up the data-set. The remaining photos are distributed across the training batches in random order, however certain training batches can have a disproportionate number of images from a particular class.The training blocks are made up of a total of 5000 images from every class. Access public database \href{https://www.cs.toronto.edu/~kriz/cifar.html}{here}
\subsubsection{CelebA}
More than 200K celebrity photos with 40 feature annotations each make up the large-scale face attributes collection known as Celeb Faces Attributes collection\cite{zhang2019show}. This collection of photos includes a wide range of poses and cluttered backgrounds. With 10,177 distinct identities, 202,599 face photos, 5 landmark locations, and 40 binary attribute annotations per image. Access public database \href{https://www.tensorflow.org/datasets/catalog/celeb_a}{here}

\subsubsection{ADE20K}
ADE20K is made up of more than 27K images from the SUN and Locations databases. The photos have been meticulously annotated with over 3K different images. Numerous images also show individual parts and pieces of other sections. Register and access the database \href{http://groups.csail.mit.edu/vision/datasets/ADE20K/request_data/}{here}

\subsection{Models and Checkpoints}
In our tests, we only utilize vision transformer (ViT) modeling \cite{dosovitskiy2021image}. The models that we optimize include four common sizes: ViT-Hybrid/S/L/B,. These models, which range in size from the smallest to the largest model by a factor of twenty eight. We take into account three distinct spatial resolutions for the picture resolutions: 476 pixels, 448 pixels and 226 pixels.

\begin{figure}
    \caption{Consequences of the validation test accuracy seen in the test with no fine-tuning
}
    \includegraphics[width=15cm,left]{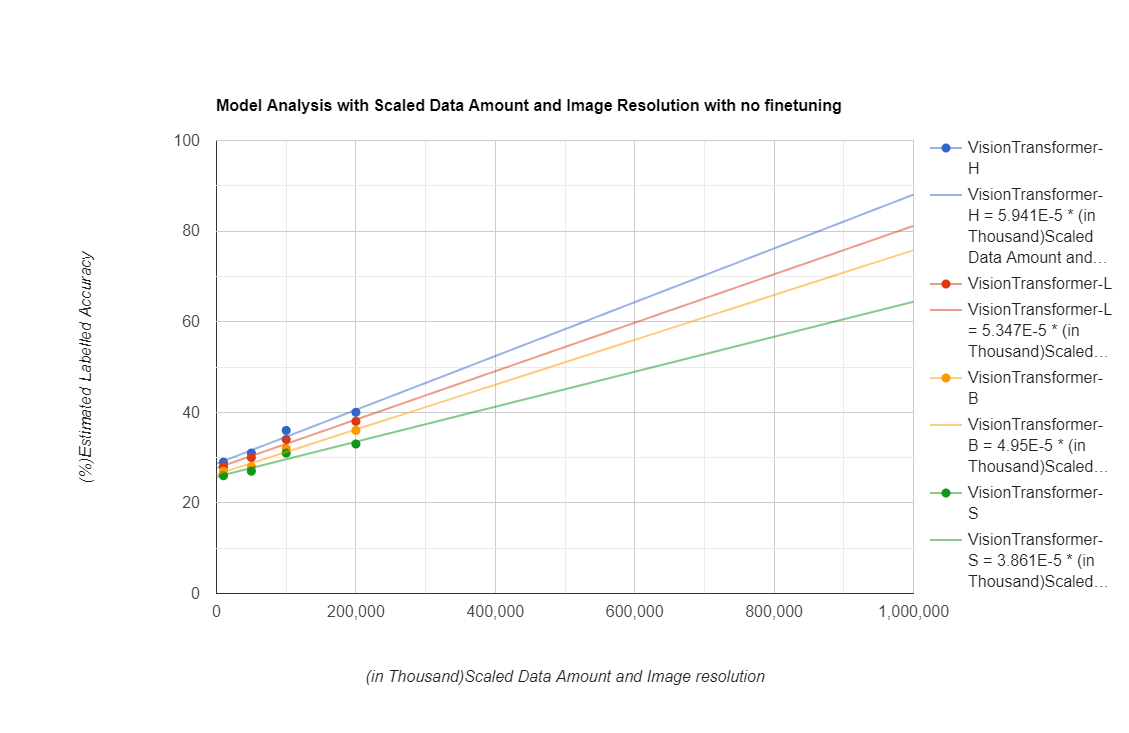}

    \label{fig:fig2}
\end{figure}

\begin{figure}
 \caption{
  Consequences of the validation test accuracy seen in the test with two percent fine-tuning
}
    \includegraphics[width=15cm,left]{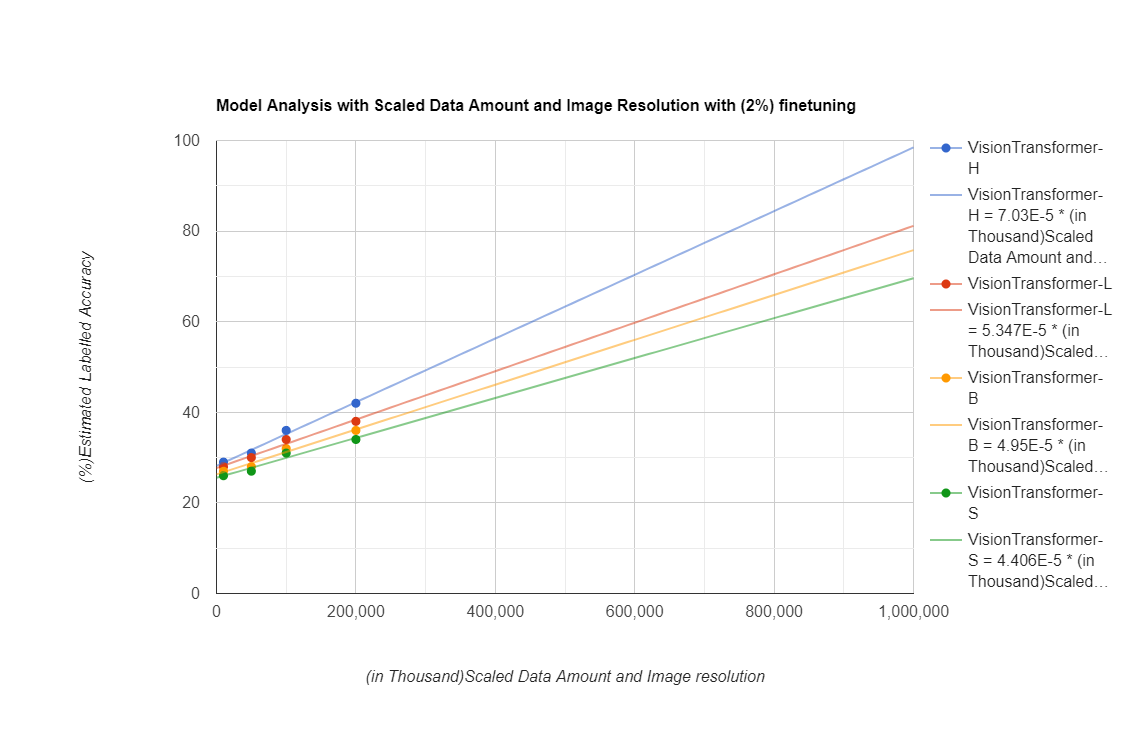}
   
    \label{fig:fig3}
\end{figure}

Our preferred Self Supervised learning method is masked auto-encoders. For the objectives of this work, masked auto-encoders offer a number of benefits against conventional self-supervised pictorial representational training methods. They first require relatively little data augmentation, in contrast to the majority of existing visual Self Supervised methods.  This is helpful for our objectives because other Self Supervised learning algorithms' significant data augmentations reduce the human-likeness of the training data. Second, we select a masking proportion of Eighty percent since masked auto-encoders perform well with such large masking proportions. 
According to the quantity of  pictures utilized for self-supervised pre-training, estimated reliability of validation on Image-Net. Performance grater than ninety percent denotes performance at the level of a human. Four distinct models are represented by various colors in the tale. The matches to the first equation are shown by the solid lines. The expected precision under four fictitious situations are shown by the lines of best fits,for additional details on these instances, see Column one . Findings for the more lenient modifying control are shown in the figure \ref{fig:fig3}, while those for the stricter fine-tuning control models without fine-tuning are shown in figure \ref{fig:fig2}.\\
Figure \ref{fig:fig2} illustrates the findings of our Image Network scalability research. We use an intuitive polynomial function in logarithmic order to represent the impact of data size, model size, and picture quality on image recognition precision:

\begin{equation}
Precision= (\beta_{i}+\log _{i}\alpha_{i})(\log _{ppi}\alpha_{ppi}+\beta_{ppi})
\end{equation}
where i is the scaled data amount which is measured in thousands here and ppi is the scaled image resolution per test.

Table \ref{tab:table} presents three possible outcomes based on the following situations: (i) a practical reference circumstance matching to our largest and best-performing model thus far, the VisionTransformer-H model, that was constructed using all of the approximately 200000 images from the public data.

A possible circumstance where we quadruple \ref{tab:table} each of i, ppi, with respect to the benchmark situation; a possible scenario  where we quadrupled \ref{tab:table} each of i, ppi with regard to the benchmark case. The image recognition rate for every one of these situations is shown in Table \ref{tab:table} as the combination of real and predicted accuracy.

The anticipated precision now surpasses the ninety percent accuracy we have defined as the lower limit on human-level precision on validation in the estimated scenario and under the additional lenient finetuning condition. This result is very encouraging because it only calls for a 522B variable Vision Transformer model trained with around 200000 images of public data. The increases in data amount and image resolution needed under this scenario are also relatively small.
\begin{table}
\centering
 \caption{Under a real-world reference situation, evaluation scaled data amount and picture resolution, and two forecasts based on adaptations with reference to \ref{fig:fig2} and \ref{fig:fig3}}
  \centering
  \begin{tabular}{lllll}
    \toprule                   
    
    Estimation Rate& Data Amount(Thousands) & i (parameters)  & Validation Accuracy (\%) & Data Resolution(ppi) \\
    \midrule
    Current Test & 200 & 0.3B & 40 & 256     \\
    4 x Test     & 2400 & 9.6B & 65 & 1280      \\
    10 x Test     & 12600& 522B & 91 & 4096  \\
    \bottomrule
  \end{tabular}
  \label{tab:table}
\end{table}

\begin{figure}
 \caption{
  Consequences of the validation with human benchmark with no fine-tuning
}
    \includegraphics[width=15cm,left]{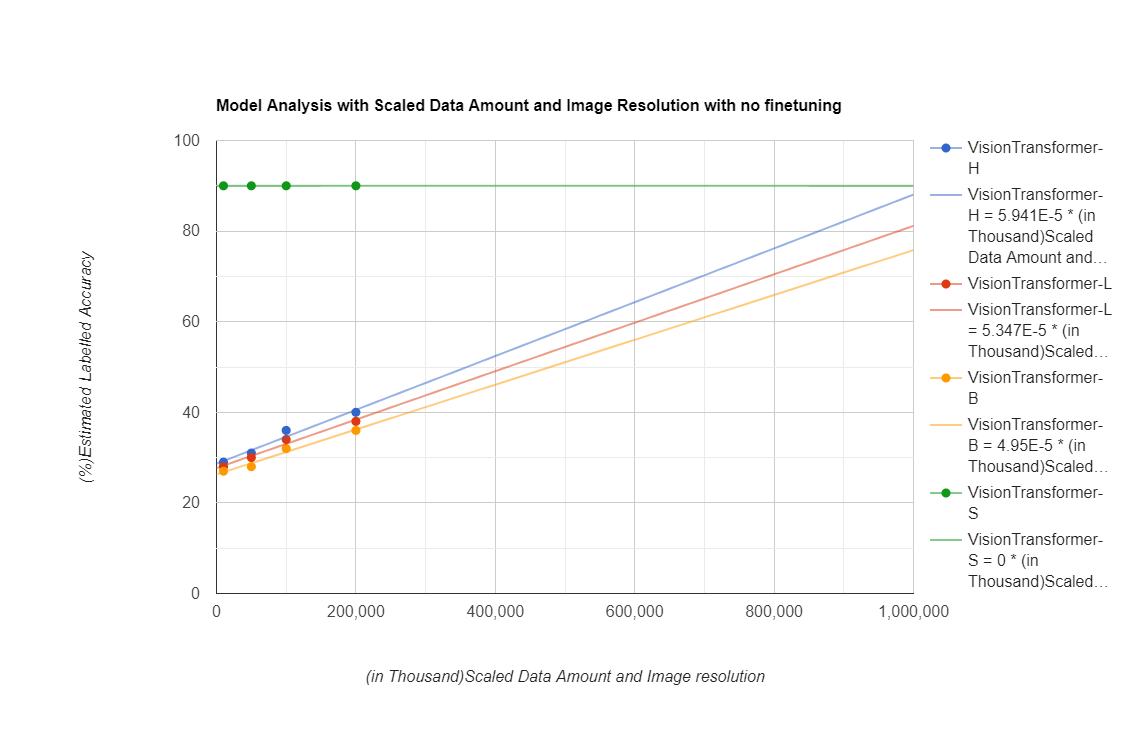}
   
    \label{fig:fig4}
\end{figure}
\begin{figure}
 \caption{
  Consequences of the validation with human benchmark with two percent fine-tuning
}
    \includegraphics[width=15cm,left]{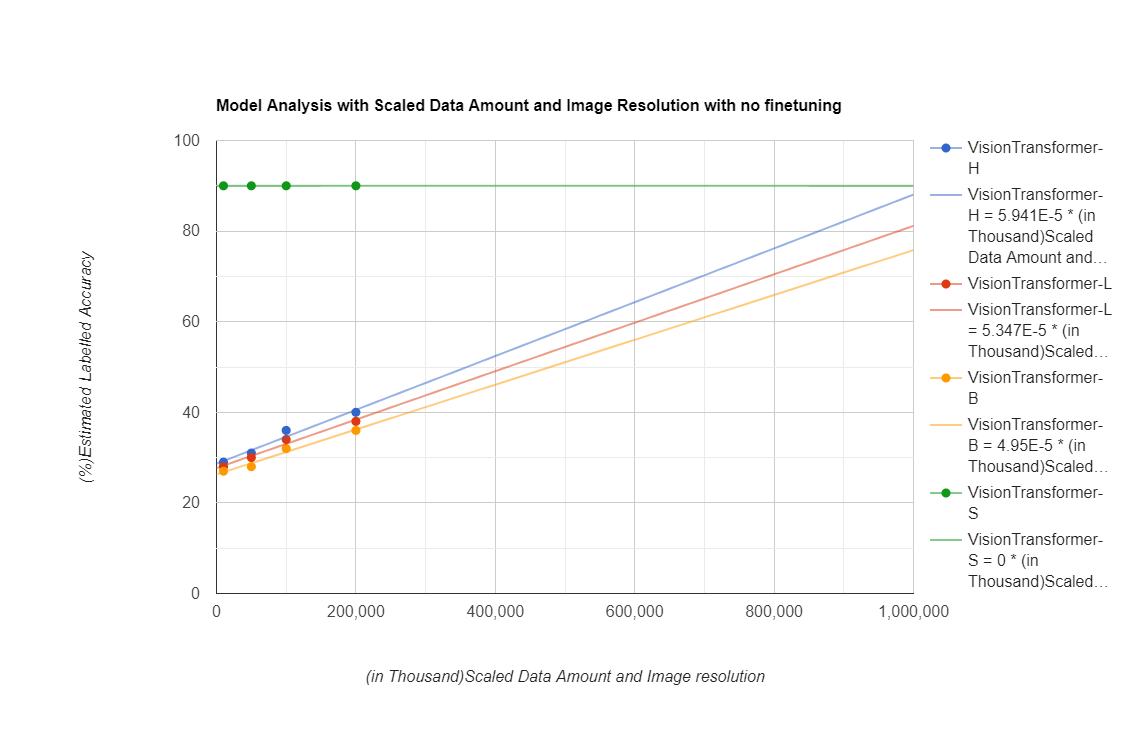}
   
    \label{fig:fig5}
\end{figure}

\section{Conclusion}
Using highly general self-supervised machine learning techniques and deep learning architectures without significant inductive biases, our results show that human-level precision as well as resilience in visual image recognition can be achieved through anthropomorphic visual perception at sub-human scales of data amount and image quality.

\section*{Acknowledgments}
This was was supported in part by Microsoft with their Azure Credits porgram

\bibliographystyle{apalike}  
\bibliography{references}

\end{document}